\pdfoutput=1
\documentclass[11pt]{article}
\usepackage[preprint]{acl}
\usepackage{times}
\usepackage{latexsym}

\usepackage[normalem]{ulem}
\usepackage{mathrsfs}
\usepackage{amsthm}
\usepackage{amsmath}
\usepackage{amssymb, bm}
\usepackage{multirow}
\usepackage{booktabs}
\usepackage{enumitem}
\usepackage{times}
\usepackage{latexsym}
\usepackage{array}
\usepackage{subfigure} %
\usepackage{multirow}
\usepackage{graphicx}
\usepackage{amsfonts}
\usepackage{subcaption}
\usepackage{xspace}
\usepackage{tabularx}
\usepackage{algorithm}
\usepackage{arydshln}
\usepackage{algpseudocode}
\usepackage{colortbl}
\usepackage{tcolorbox}
\usepackage{array}  %% 사용해도 되는지 확인해보십쇼! 
\usepackage{soul}
\usepackage[T1]{fontenc}
\usepackage[utf8]{inputenc}
\usepackage{microtype}
\usepackage{inconsolata}
\renewcommand*{\thefootnote}{\fnsymbol{footnote}}
\title{Crafting the Path: Robust Query Rewriting for Information Retrieval}

\author{Ingeol Baek,~Jimin Lee,~Joonho Yang,~Hwanhee Lee\textsuperscript{$\dagger$}\\
{Department of Artificial Intelligence, Chung-Ang University, Seoul, Korea} \\
\texttt{baekingeol98@gmail.com}, \quad\texttt{\{ljm1690, plm3332, hwanheelee\}@cau.ac.kr} \\
}

\newcommand{\name}{\textsc{Crafting The Path}}

\begin{document}
\maketitle
\footnotetext{\textsuperscript{$\dagger$}Corresponding author.}
\renewcommand*{\thefootnote}{\arabic{footnote}}
\begin{abstract}
Query rewriting aims to generate a new query that can complement the original query to improve the information retrieval system. Recent studies on query rewriting, such as \textit{query2doc}, \textit{query2expand} and \textit{querey2cot}, rely on the internal knowledge of Large Language Models (LLMs) to generate a relevant passage to add information to the query. Nevertheless, the efficacy of these methodologies may markedly decline in instances where the requisite knowledge is not encapsulated within the model's intrinsic parameters. In this paper, we propose a novel structured query rewriting method called \name{} tailored for retrieval systems. \name{} involves a three-step process that crafts query-related information necessary for finding the passages to be searched in each step. Specifically, the \name{} begins with \textit{Query Concept Comprehension}, proceeds to \textit{Query Type Identification}, and finally conducts \textit{Expected Answer Extraction}. Experimental results show that our method outperforms previous rewriting methods, especially in less familiar domains for LLMs. We demonstrate that our method is less dependent on the internal parameter knowledge of the model and generates queries with fewer factual inaccuracies. Furthermore, we observe that \name{} demonstrates superior performance in the retrieval-augmented generation scenarios.

\end{abstract}

\section{Introduction}
\label{lab:intro}
% \begin{figure}[ht]
% \centerline{\includegraphics[scale=0.48]{./images/figure1.pdf}}
% \vspace{-2mm}
% \caption{Structure of query rewriting framework: The dash line indicates a case where query rewriting is performed and the answer is already available, so there is no need for an additional retriever.}
% \vspace{-7mm}
% \label{fig:retr}
% \end{figure}
In an open-domain QA system~\citep{lewis2020retrieval, zhu2021retrieving, li2022survey, zhang-etal-2023-survey-efficient, kamalloo2023evaluating}, document retrievers are utilized to retrieve the necessary information to answer the given query. Query rewriting reformulates original queries to help the retrieval system find relevant passages. Recent works on query rewriting focus on Large Language Models (LLMs)~\cite{NEURIPS2020_1457c0d6, OpenAI_ChatGPT, openai2023gpt4, touvron2023llama} to generate additional information. Specifically, these studies aim to generate a relevant passage for a given query by leveraging the pre-trained knowledge of LLMs. Utilizing these new queries generated from LLMs has shown a significant increase in the performance of retrieval systems. Recently, various LLM based rewriting methods such as \textit{query2doc} (Q2D)~\citep{wang2023query2doc}, \textit{query2expand} (Q2E), and \textit{querey2cot} (Q2C)~\citep{jagerman2023query} have been introduced.
Q2D generates a pseudo-document based on the original query, which is then used as input for the retriever. Similarly, Q2C employs a Chain-of-Thought~\citep{wei2022chain} approach, and Q2E generates semantically equivalent queries. These approaches leverage the rewriting of the original query into a form similar to passages in the corpus. These techniques result in superior performance improvements compared to the base query alone.

\begin{figure*}[!ht]
\centerline{\includegraphics[scale=0.50]{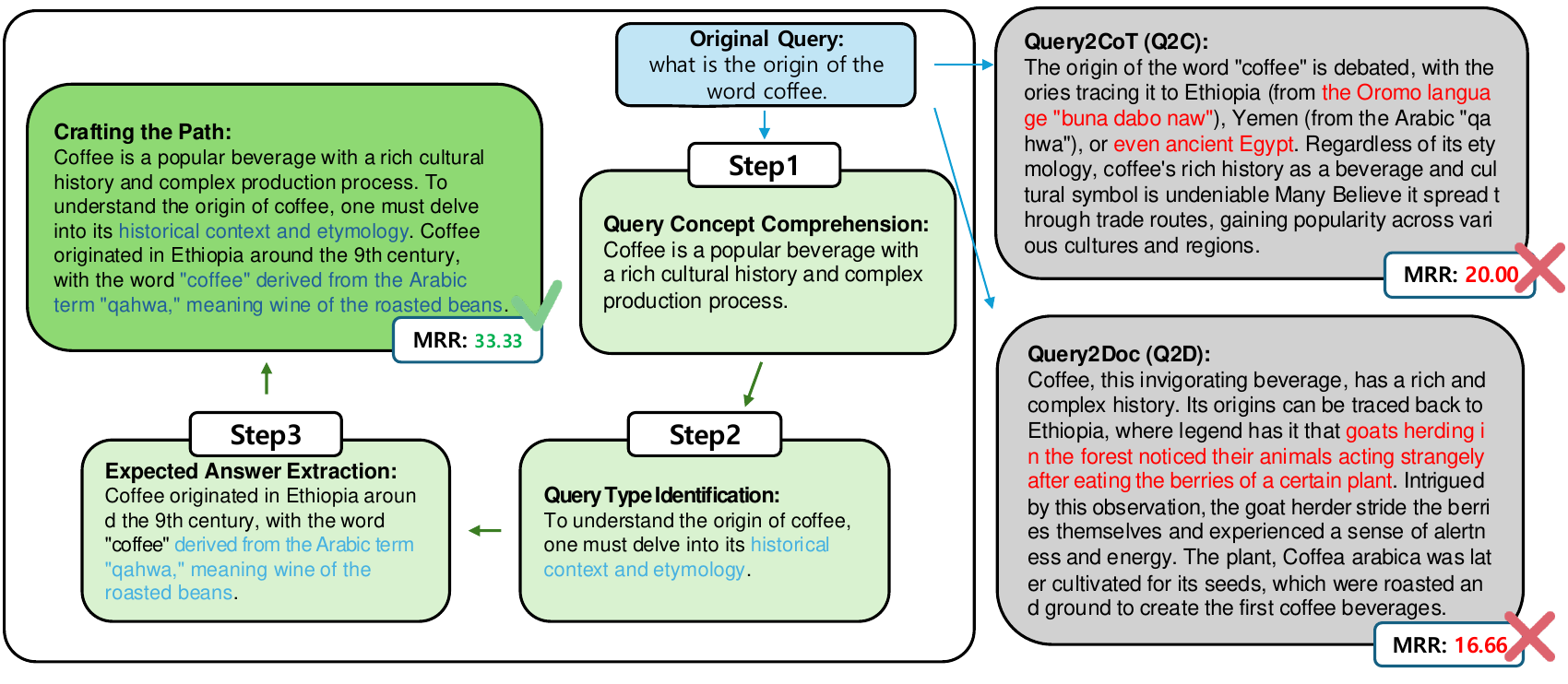}}
% \vspace{-4mm}
\caption{Overview of our proposed query rewriting method \name{}, along with the rewritten query examples of \textit{query2doc} (Q2D) and \textit{query2cot} (Q2C) methodologies. We represent the factual error in red and the accurate information in blue.}
\vspace{-5mm}
\label{fig:crafting}
\end{figure*}

% In Figure~\ref{fig:retr}, the dash line shows that the rewriting model's internal knowledge sometimes generates answers  during query rewriting. However, 
The fundamental reason for utilizing retrieval systems in open-domain QA is to use external knowledge when QA systems do not know to generate correct answer~\citep{gao2023retrieval}. And this is the original purpose of using Retrieval-Augmented Generation (RAG) systems~\citep{10.5555/3495724.3496517, wang2023survey}. 
% It is important to consider the retrieval performance changes through query rewriting in cases where the rewriting model generates an incorrect answer to the query~\citep{wang2023survey}. 
However, relying heavily on inherent knowledge of LLMs for query rewriting often leads to the generation of irrelevant information and causes numerous factual errors~\citep{chern2023factool, min2023factscore, luo2023chatgpt} in the reformulated queries. As in the right examples of Figure~\ref{fig:crafting}, Q2C asserts that coffee originated in ancient Egypt or Yemen, whereas its origin is Ethiopia. Meanwhile, Q2D discusses a legendary story about "goats" without providing any information about the origin of the word "coffee". These types of misinformation and unrelated contents can lead to significant performance degradation because of the included incorrect information in the reformulated queries.

In this paper, we propose a novel query rewriting method, \name{}, which is a fine-grained query reformulation technique through the structured reasoning process. Instead of simply generating the passage similar to the candidate documents, \name{} focuses on identifying what information needs to be found to solve the given query. The \name{} method comprises three steps. The first step, \textit{Query Concept Comprehension}, provides fundamental background knowledge. Offering basic factual information reduces the likelihood of including incorrect information and helps the retrieval system clearly understand the main topic. The second step, \textit{Query Type Identification}, specifies the required information to filter out irrelevant information. Finally, through \textit{Expected Answer Extraction} process, the retriever model identifies the essential information it needs to find, facilitating the extraction of accurate passages. This structured, process minimizes unnecessary inferences by the model, thereby reducing the possibility of factual errors. 

\name{} outperforms all of the baseline methods and exhibits a reduced degree of factual errors demonstrated by the 10\% higher FActScore~\citep{min2023factscore} compared to baselines. Additionally, our approach demonstrates enhanced performance without prior knowledge of the model’s internal parameters. 
This is evidenced by experiments in closed-book QA settings, where rewriting models fail to provide correct answers, resulting in a 3.57\% increase in retrieval performance.
Furthermore, our method shows 7.3\% less latency compared to the baselines and also demonstrates superior performance in adopting query rewriting to open-domain QA scenarios.

\section{\name{}}
\label{sec:method}
Our proposed rewriting method, \name{}, is composed of the following three steps.
\subsection{Query Rewriting via \name{}}
\subsubsection{Step 1: \textit{Query Concept Comprehension}} We begin with \textit{Query Concept Comprehension} step, which generates additional information that serves as the contextual background for the existing query. This step enriches the direct information about the question object within the query. This step performs a higher level of abstraction for the original question. As shown in Figure~\ref{fig:crafting}, it elaborates on the high-level concept and provides detailed explanations about “coffee,” which is part of the original query. \textit{Query Concept Comprehension} step plays a crucial role in generating high-level information while aiding in identifying the information to be searched for in the next step.
% For example, in Figure~\ref{fig:crafting}, when asking about the origin of "coffee," this step provides a detailed description of "coffee". 
% Query Concept Comprehension step aids in identifying the information to be searched for in the next step, also playing a crucial role in generating high-level information.Í
\subsubsection{Step 2: \textit{Query Type Identification}} 
Based on the specific information obtained through the original query and Step 1, we proceed to \textit{Query Type Identification} step. In this step, we generate the necessary information to retrieve relevant passages. Specifically, we create categories for the query that help filter out irrelevant information. To retrieve information to answer the origin of coffee as in Figure~\ref{fig:crafting}, we can think that one must search for the historical context and etymology of coffee. This step can filter out irrelevant passages because it specifies the information to be found through rewriting, which serves as input for the retriever model. Inspired by this point, \textit{Query Type Identification} aims to find the type of necessary information that the ground truth passage might include as in Figure~\ref{fig:crafting}, and it helps to identify passages containing information on the “historical context and etymology” of coffee.

\subsubsection{Step 3: \textit{Expected Answer Extraction}} The final step involves extracting expected answers for the query based on the information generated from the previous step. 
As in Figure~\ref{fig:crafting}, the details regarding the origin of coffee being Ethiopia and its etymology enable the retriever model to identify the required information, facilitating the extraction of accurate passage. 
% In Figure~\ref{fig:crafting}, Step 2 states that based on the historical context and etymology mentioned, the word coffee is derived from the 9th-century Ethiopian term "qahwa".
\begin{table}[ht]
  \small
  \centering
  \begin{tabularx}{\linewidth}{X}
    \toprule
    \textbf{Input Prompt for \name{}} \\
    \midrule
\textbf{Instruction:}
By following the requirements, write 3 steps related to the Query and answer in the same format as the example.
\\
\\
\textbf{Requirements:}\\
1. In step1, generate the contextual background from the existing query is extracted.\\
2. In step2, generate what information is needed to solve the question.\\
3. In step3, generate expected answer based on query, step1, and step2.\\
4. If you think there is no more suitable answer, end up with 'None'.\\
\\
\textbf{Query 1:} what is the number one formula one car? \\
\textbf{Step 1:} Formula One (F1) is the highest class of international automobile racing competition held by the FIA.\\
\textbf{Step 2:} To know the best car, you have to look at the race records.\\
\textbf{Step 3:} Red Bull Racing's RB20 is the best car.\\
\\
   (4-shot examples) ...\\
\\
\textbf{Query 5:}\\
    \bottomrule
  \end{tabularx}
  
\caption{Prompt used for \name{}.}
\vspace{-4mm}
\label{tab:ctp_prompt}
\end{table}

These three distinct steps offer a form of query rewriting that enhances the retrieval of more accurate information and minimizes the inclusion of incorrect information, resulting in better Open-domain QA performance in Figure~\ref{fig:qa}. We implement all of these steps in \name{} with a single LLM call using the prompt in Table \ref{tab:ctp_prompt}. Specifically, we provide the role of each step with the examples in the prompts. Additionally, to avoid producing inaccurate information, we instruct the model to generate \textit{“None”} when it lacks certain knowledge, providing clear guidance for the retriever system’s input. 

\subsection{Passage Retriever}
\textbf{Constructing Inputs of Retriever} To construct the final query $q^+$, we expand the original query $q$ three times and concatenate $q$ with rewritten query $QR$ in sparse retrieval as shown in Eq. \ref{eq:sparse}. In the case of dense retrieval, a [SEP] token is inserted between the query and the QR to differentiate them. 
\begin{equation} 
\resizebox{.7\hsize}{!}{
    $\text{Sparse: } q^+ = \text{concat(}\{q\} \times \text{3},\ QR\text{)}$.
    }
\label{eq:sparse}
\end{equation} 
\begin{equation} 
\resizebox{.7\hsize}{!}{
    $\text{Dense: } q^+ = \text{concat(}q,\text{[SEP]}, \ QR\text{)}$.
    }
\label{eq:dense}
\end{equation} 

\textbf{Training Dense Retriever} To train a dense retriever, we utilize the Binary Passage Retrieval (BPR) loss~\citep{yamada-etal-2021-efficient} as follows to reduce the memory usage:
\begin{equation}
  \resizebox{.89\hsize}{!}{
  $\mathcal{L}_{\text{cand}} = \sum_{j=1}^{n}\max(0, -(\langle \mathbf{\tilde{h}}_{q_i^{\phantom{0}}}, \mathbf{\tilde{h}}_{p_i^+}\rangle + \langle \mathbf{\tilde{h}}_{q_i^{\phantom{0}}}, \mathbf{\tilde{h}}_{p_{i,j}^-}\rangle) + \alpha)$.
  }
  \label{eq:ranking-loss}
\end{equation}
\begin{equation}
  \resizebox{.89\hsize}{!}{
  $\mathcal{L}_{\text{rerank}} = -\log \frac{\exp(\langle \mathbf{e}_{q_i^{\phantom{0}}},\, \mathbf{\tilde{h}}_{p_i^+}\rangle)}{\exp(\langle \mathbf{e}_{q^{\phantom{0}}_i},\, \mathbf{\tilde{h}}_{p_i^+}\rangle) + \sum_{j=1}^n{\exp(\langle \mathbf{e}_{q^{\phantom{0}}_i},\, \mathbf{\tilde{h}}_{p^-_{i,j}}\rangle)}}$
  },
  \label{eq:reranking-loss}
\end{equation}
where $\mathcal{D} = \{ \langle q_i, p^+_i, p^-_{i,1}, \cdots, p^-_{i,n} \rangle \}_{i=1}^m$ denote a set where $m$ represents training instances, $p^+_i$ denotes a positive passage, and $p^-_{i,j}$ denotes a negative passage. We compute embedding $e \in \mathbb{R}^d$ using an encoder, each $\tilde{h}_q$ and $\tilde{h}_p$ represent the hash code for a query and a passage, respectively.
$\mathcal{L}_{cand}$ is to identify positive passages based on ranking loss, and $\alpha$ is the margin that is enforced between the positive and negative scores. 
$\mathcal{L}_{rerank}$ is used to minimize the negative log-likelihood for a positive passage. Finally, we employ the BPR loss as follows:
\begin{equation}
    \mathcal{L}_{bpr} = \mathcal{L}_{cand} + \mathcal{L}_{rerank}.
\end{equation}
\section{Experiments}
\begin{table*}[ht]

\renewcommand{\arraystretch}{1.4}

\resizebox{\textwidth}{!}{
\begin{tabular}{llccccccccc||c}

\hline
\multicolumn{2}{c}{\begin{tabular}[c]{@{}c@{}}\end{tabular}}&scifact&trec-covid&nfcorpus&quora&scidocs&hotpotqa&dbpedia&fiqa&\multicolumn{1}{c}{fever}&\multicolumn{1}{c}{Avg}\\\hline
&Ours&{58.41$^\text{(±0.10)}$}&{\textbf{64.59}$^{\text{(±0.86)}}$}&{\textbf{32.28}$^\text{(±0.03)}$}&{\textbf{75.21}$^{\text{(±2.27)}}$}&\ul{18.39}$^\text{(±0.10)}$&{50.66$^{\text{(±0.36)}}$}&{\textbf{41.78}$^{\text{(±0.18)}}$}&\ul{40.74}$^\text{(±0.09)}$&\ul{63.21}$^{\text{(±0.26)}}$&{\color[HTML]{FF0000}\textbf{49.47}}\\

Mistral-7b&Q2D&{\textbf{59.40}$^{\text{(±0.07)}}$}&{61.92$^{\text{(±1.07)}}$}&{32.03$^\text{(±0.07)}$}&\ul{74.79}$^{\text{(±0.41)}}$&{\textbf{18.40}$^\text{(±0.03)}$}&\ul{52.05}$^{\text{(±0.59)}}$&{41.46$^{\text{(±0.48)}}$}&{40.37$^\text{(±0.41)}$}&{\textbf{64.26}$^{\text{(±0.60)}}$}&\ul{49.41}\\

{Dense$^{\text{FT}}$}&Q2E&{57.06$^\text{(±2.11)}$}&{55.89$^\text{(±2.38)}$}&{31.98$^\text{(±0.23)}$}&{71.98$^\text{(±0.98)}$}&{18.29$^\text{(±0.01)}$}&{48.57$^\text{(±0.50)}$}&{39.13$^\text{(±0.23)}$}&{\textbf{40.90}$^\text{(±0.71)}$}&{60.85$^\text{(±1.20)}$}&{\color[HTML]{333333}47.19}\\

&Q2C&\ul{59.37}$^\text{(±1.83)}$&\ul{63.14}$^{\text{(±6.22)}}$&\ul{32.15}$^\text{(±0.16)}$&{74.33$^{\text{(±2.43)}}$}&{18.31$^\text{(±0.05)}$}&{\textbf{52.27}$^{\text{(±1.18)}}$}&\ul{41.75}$^{\text{(±0.14)}}$&{40.60$^\text{(±0.31)}$}&{62.60$^{\text{(±0.48)}}$}&{\color[HTML]{333333}49.39}\\\hline
% &Ours&\ul{55.28}$^\text{(±0.02)}$&{\textbf{85.16}$^\text{(±7.28)}$}&{\textbf{53.42}$^\text{(±0.60)}$}&{\textbf{73.65}$^\text{(±2.30)}$}&{\textbf{32.53}$^\text{(±0.43)}$}&{67.94$^\text{(±0.45)}$}&{\textbf{71.57}$^\text{(±1.58)}$}&\ul{48.61}$^\text{(±0.36)}$&\ul{62.46}$^\text{(±0.42)}$&{\color[HTML]{FF0000}\textbf{61.18}}\\
% &Q2D&{55.81$^\text{(±0.11)}$}&{81.55$^\text{(±6.04)}$}&{53.00$^\text{(±0.24)}$}&\ul{73.22}$^\text{(±0.47)}$&\ul{32.13}$^\text{(±0.05)}$&\ul{69.10}$^\text{(±0.78)}$&\ul{71.49}$^\text{(±1.85)}$&{47.93$^\text{(±0.38)}$}&{\textbf{63.55}$^\text{(±0.67)}$}&60.87\\
% &Q2E&{53.89$^\text{(±2.18)}$}&{75.99$^\text{(±2.20)}$}&{52.44$^\text{(±0.67)}$}&{70.62$^\text{(±1.06)}$}&{31.81$^\text{(±0.13)}$}&{65.51$^\text{(±0.82)}$}&{65.55$^\text{(±4.68)}$}&{\textbf{48.66}$^\text{(±0.96)}$}&{59.68$^\text{(±1.47)}$}&58.24\\
% \multirow{-4}{*}{MRR}&Q2C&{\textbf{56.30}$^\text{(±1.36)}$}&\ul{84.04}$^\text{(±6.92)}$&\ul{53.03}$^\text{(±0.33)}$&{72.75$^\text{(±2.31)}$}&{31.86$^\text{(±0.15)}$}&{\textbf{69.24}$^\text{(±1.36)}$}&{71.35$^\text{(±1.17)}$}&{48.23$^\text{(±0.45)}$}&{61.69$^\text{(±0.48)}$}&\ul{60.94}\\\hline

&Ours&{\textbf{58.40}$^\text{(±0.09)}$}&{\textbf{63.24}$^{\text{(±1.70)}}$}&{\textbf{32.60}$^{\text{(±0.02)}}$}&\ul{75.44}$^{\text{(±1.96)}}$&{17.99$^\text{(±0.06)}$}&{\textbf{49.84}$^{\text{(±0.23)}}$}&{\textbf{40.32}$^{\text{(±0.20)}}$}&\ul{40.31}$^{\text{(±0.09)}}$&\ul{62.12}$^{\text{(±0.39)}}$&{\color[HTML]{FF0000}\textbf{48.92}}\\
Phi-2&Q2D&\ul{58.23}$^\text{(±0.29)}$&{59.50$^{\text{(±0.68)}}$}&{31.52$^\text{(±0.13)}$}&{\textbf{75.55}$^{\text{(±0.43)}}$}&\ul{18.05}$^\text{(±0.01)}$&{49.28$^{\text{(±0.80)}}$}&{39.68$^{\text{(±0.71)}}$}&{39.81$^\text{(±0.04)}$}&{\textbf{62.73}$^{\text{(±1.01)}}$}&\ul{48.26}\\
{Dense$^{\text{FT}}$}&Q2E&{57.07$^\text{(±3.23)}$}&{56.58$^\text{(±2.82)}$}&{31.47$^\text{(±0.16)}$}&{72.46$^\text{(±1.45)}$}&{\textbf{18.26}$^\text{(±0.03)}$}&{46.94$^\text{(±0.43)}$}&{38.44$^\text{(±0.08)}$}&{\textbf{40.61}$^\text{(±0.53)}$}&{59.84$^\text{(±1.26)}$}&{46.85}\\
&Q2C&{57.84$^\text{(±1.36)}$}&\ul{62.23}$^{\text{(±4.59)}}$&\ul{32.20}$^\text{(±0.17)}$&{73.56$^\text{(±4.99)}$}&{18.00$^\text{(±0.02)}$}&\ul{49.39}$^{\text{(±0.89)}}$&\ul{40.20}$^{\text{(±0.16)}}$&{39.61$^\text{(±0.45)}$}&{60.95$^\text{(±0.27)}$}&{48.22}\\\hline
% &Ours&{\textbf{55.47}$^\text{(±0.18)}$}&{\textbf{82.70}$^\text{(±6.63)}$}&{\textbf{53.76}$^\text{(±0.11)}$}&{\textbf{73.93}$^\text{(±2.06)}$}&{\textbf{31.78}$^\text{(±0.20)}$}&{\textbf{66.98}$^\text{(±0.32)}$}&\ul{68.27}$^\text{(±0.64)}$&\ul{48.21}$^\text{(±0.22)}$&\ul{61.22}$^\text{(±0.60)}$&{\color[HTML]{FF0000}\textbf{60.26}}\\
% &Q2D&{55.15$^\text{(±0.35)}$}&{79.66$^\text{(±5.60)}$}&{51.42$^\text{(±0.24)}$}&\ul{73.91}$^\text{(±0.49)}$&\ul{31.57}$^\text{(±0.02)}$&\ul{66.48}$^\text{(±1.34)}$&{67.87$^\text{(±0.58)}$}&{47.46$^\text{(±0.15)}$}&{\textbf{61.67}$^\text{(±1.23)}$}&{59.46}\\
% &Q2E&{53.91$^\text{(±3.24)}$}&{78.07$^\text{(±3.92)}$}&{51.42$^\text{(±0.35)}$}&{71.16$^\text{(±1.43)}$}&{31.71$^\text{(±0.08)}$}&{63.82$^\text{(±0.68)}$}&{65.01$^\text{(±2.08)}$}&{\textbf{48.29}$^\text{(±0.75)}$}&{58.69$^\text{(±1.61)}$}&{58.01}\\
% \multirow{-4}{*}{MRR}&Q2C&\ul{55.21}$^\text{(±0.90)}$&\ul{82.58}$^\text{(±17.39)}$&\ul{53.15}$^\text{(±0.25)}$&{72.00$^\text{(±4.81)}$}&{31.38$^\text{(±0.12)}$}&{66.41$^\text{(±1.18)}$}&{\textbf{69.86}$^\text{(±2.46)}$}&{47.33$^\text{(±0.51)}$}&{59.81$^\text{(±0.22)}$}&\ul{59.75}\\\hline
\hline
&Ours&70.78&{\textbf{74.79}}&{\textbf{35.51}}&\ul{75.82}&\ul{16.10}&{57.41}&{\textbf{46.68}}&\ul{29.12}&{63.66}&{\color[HTML]{FF0000}\textbf{52.21}}\\
Mistral-7b&Q2D&\ul{71.14}&{67.73}&{35.01}&{70.37}&{15.68}&{\textbf{58.82}}&{41.36}&{28.67}&{\textbf{67.72}}&{50.72}\\
Sparse (BM25)&Q2E&{68.61}&{66.29}&{35.14}&{\textbf{76.66}}&{\textbf{16.13}}&{54.58}&{42.69}&{28.15}&{52.98}&{49.03}\\
&Q2C&{\textbf{71.63}}&\ul{74.45}&\ul{35.29}&{74.86}&\ul{16.10}&\ul{58.71}&\ul{43.16}&{\textbf{30.91}}&\ul{64.36}&\ul{52.16}\\
\hline
&Ours&{68.48}&{\textbf{69.71}}&\ul{34.38}&{74.47}&{15.47}&\ul{55.20}&{\textbf{42.93}}&{\textbf{28.91}}&\ul{59.20}&{\color[HTML]{FF0000}\textbf{49.86}}\\
Phi-2&Q2D&{67.15}&{59.42}&{32.83}&{\color[HTML]{000000}71.45}&{14.95}&{52.61}&{37.50}&{25.10}&{57.67}&{46.52}\\
Sparse (BM25)&Q2E&\ul{68.71}&{66.94}&{33.51}&\textbf{77.35}&{\textbf{16.21}}&{53.35}&{39.18}&{26.72}&{52.53}&{48.28}\\
&Q2C&{\textbf{69.14}}&\ul{69.57}&{\textbf{35.21}}&\ul{75.33}&\ul{15.70}&{\textbf{55.39}}&\ul{39.64}&\ul{28.37}&{\textbf{59.57}}&\ul{49.77}\\\hline
\end{tabular}}
\vspace{-1mm}
\caption{Experimental results on BEIR dataset. Highest performance is highlighted in \textbf{bold}, and the second highest is \ul{underlined}.}
%\vspace{-4mm}
\label{tab:ood}
\end{table*}

%  The model, trained with the MS-MARCO Passage dataset, was directly utilized as a retrieval model. Experiments were performed using nine selected datasets from BEIR. For each rewriting method, five trials were conducted to calculate the mean and variance. \blacklozenge
\subsection{Experimental Setup}
\paragraph{Datasets}
We use the MS-MARCO passage dataset~\citep{Campos2016MSMA} for training retriever. Additionally, To demonstrate the robustness of our model on unseen data, we utilize nine retrieval datasets from BEIR~\citep{thakur2021beir} for our main experiment. In our experiment, we utilize the nDCG@10 metric to evaluate the quality of the top 10 search results based on their relevance and order. Additionally, we use the HotpotQA~\citep{yang2018hotpotqa} and NaturalQA~\citep{47761} datasets for the open-domain QA experiment and measure the accuracy.\\
%which is a statistical measure used to evaluate the effectiveness of query retrieval systems by calculating the average inverse rank of the first correct answer.\\

\paragraph{Baselines}
To analyze query rewriting methods based on LLMs, we use three baselines: \textit{query2doc} (Q2D), \textit{query2cot} (Q2C), and \textit{query2expand} (Q2E). All rewriting methods use 4-shot prompts. For Q2D, Q2E and Q2C, we reference the prompt from~\citet{wang2023query2doc}, and~\citet{jagerman2023query}. 
% \name{} involves writing content for each step and progressively generating new queries. 
In our experiments, we employ Mistral-7b~\citep{jiang2023mistral} and Phi-2~\citep{Phi-2} as query rewriting models including \name{}. We conduct both dense and sparse retrieval experiments with all rewriting methods on the 9 BEIR datasets.\\
\paragraph{Implementation Details} %We train all of the dense retrieval models using the rewritten queries produced by Mistral-7b. 
For reliable experiments, we train five different retriever models using a different seed for each method. To evaluate the query rewriting methods using dense retrieval models, we use a total of 20 models. For each dataset, we construct 5 (\textit{fine-tuned retrieval models}) $\times$ 4 (\textit{rewriting methods}) = 20 embedding vectors, and compute the mean and variance.
% To evaluate query rewriting methods with the dense retriever models, we experiment 5 (\textit{fine-tuned retriever model}) $\times$ 4 (\textit{rewriting methods}) = 20 trials to get statistically reliable results. % for each dataset using a total of 20 models, and compute the mean and variance. 
In experiments with the query rewriting method using Phi-2, we obtain results using models trained with the new queries written by Mistral-7b. We employ all-mpnet-base-v2\footnote{https://huggingface.co/sentence-transformers/all-mpnet-base-v2} as our pre-trained dense retrieval model. For the sparse retriever system, We use the default BM25 parameters provided by pyterrier~\citep{pyterrier2020ictir}\footnote{https://pyterrier.readthedocs.io/}.

\begin{figure*}[!ht]
\centerline{\includegraphics[scale=0.45]{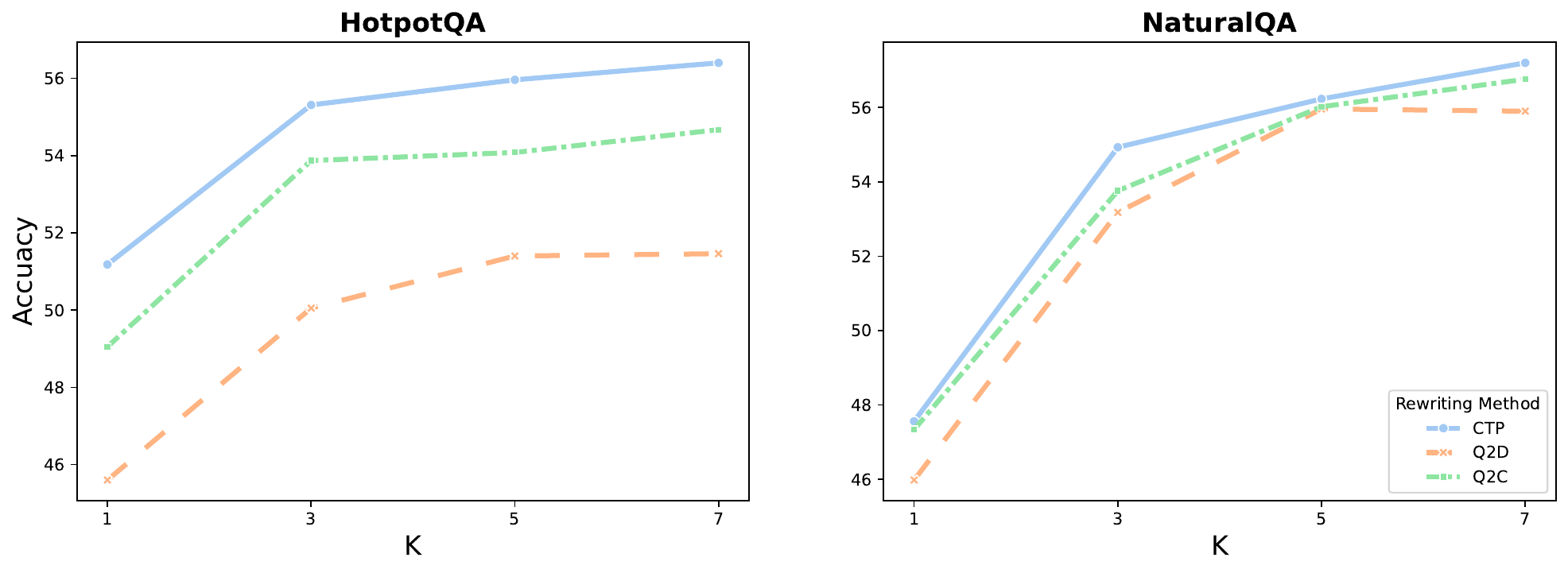}}
% \vspace{-4mm}
\caption{The Retrieval-Augmented Generation performance of HotpotQA (left) and NaturalQA (right), when performing \name{} (CTP), \textit{query2doc} (Q2D), and \textit{query2cot} (Q2C). K means the number of retrieved passages.}
% \vspace{-6mm}
\label{fig:qa}
\end{figure*}

\subsection{Results}
\paragraph{Main Results}
As shown in Table~\ref{tab:ood}, \name{} outperforms all approaches, including query2doc (Q2D), query2cot (Q2C), and query2expand (Q2E), for the average score across 9 BEIR datasets. Our approach consistently demonstrates superior performance compared to existing methods. \name{} provides a robust application across various domains and rewriting model sizes in both sparse and dense retrievers. Moreover, we observe significant performance improvements on the trec-covid~\citep{Wang2020Cord19,Voorhees2020TrecCovid} and nfcorpus~\citep{Boteva2016Nfcorpus} datasets for both retriever types compared to previous methods. Especially, trec-covid dataset requires searching for the latest information on queries about COVID-19. \name{} proves to be more effective in finding such recent information. However, previous methods outperform our approach on datasets like FEVER~\citep{thorne-etal-2018-fever}. The FEVER dataset is based on Wikipedia, which is frequently used for LLMs pre-training data. Since the training data for the Mistral-7b and Phi-2 models are not disclosed, and verifying the presence of internal parameter knowledge directly remains a challenge, it is difficult to confirm the internal knowledge of the models~\citep{wang2023survey}. However, we conduct measurements of the impact of model's internal parameter knowledge through experiments in Table~\ref{tab:closedbook}. %However, we experimentally demonstrate that our approach robustly applies not only to general tasks but also to tasks that require the retrieval of the latest information. 
As results in Table~\ref{tab:ood}, we observe the significant performance improvements in these BEIR datasets from \name{} rewriting methods through the use of a structured rewriting method.

\paragraph{Ablation Study}
% \begin{table}[]
% \small
% \centering
% \setlength{\tabcolsep}{3pt}
% \renewcommand{\arraystretch}{1.15}
% \setlength{\abovecaptionskip}{0.3cm}
% \setlength{\belowcaptionskip}{-0.0cm}
% \begin{tabular}{lccccc}
% \hline
%                           & \multicolumn{5}{c}{MS-MARCO Passage dev}          \\
%                           & nDCG  & MRR   & Recall@10 & Recall@50 & Recall@1K \\ \hline
% Ours (w/o step2, 3) & 44.59 & 32.42 & 57.24     & 78.90     & 96.17     \\
% Ours (w/o step3)    & 44.99 & 32.91 & 58.27     & 79.52     & 96.67     \\
% Ours                & 45.42 & 33.11 & 58.89     & 80.59     & 97.05     \\ \hline
% \end{tabular}
% \label{tab:ablation}
% \caption{Ablation study}
% \end{table}

\begin{table}[t]
% \tiny
\small
\centering
\renewcommand{\arraystretch}{1.3}
\resizebox{\linewidth}{!}{
\begin{tabular}{lccc}
\hline
                    & \multicolumn{3}{c}{MS-MARCO Passage dev} \\
                    & nDCG       & MRR        & Recall@1K      \\ \hline
\name{}              & 45.42      & 33.11      & 97.05          \\
\quad \textit{w/o step3}    & 44.99      & 32.91      & 96.67          \\
\quad \textit{w/o step2, 3} & 44.59      & 32.42      & 96.17          \\ \hline
\end{tabular}
}
\caption{Ablation study on \name{} method.}
\vspace{-4mm}
\label{tab:ablation}
\end{table}
We conduct an ablation study on the \name{} method to observe the influence of each step. We measure the performance on the MS-MARCO dataset by excluding each step as shown in Table~\ref{tab:ablation}. Also, We utilize nDCG, MRR, and Recall@1K to measure performance. Our experiment with omitting Step 3 and also conduct experiments excluding both Steps 2 and 3. Performance declines with the removal of each step, demonstrating that each step is essential for performance improvement.

\subsection{Analysis} 
\label{sec:analysis}
\paragraph{Measuring the Reliance on Internal Knowledge}
To evaluate the reliance of the internal model parameter knowledge in query reformulation of each LLM, we divide the dataset into problems where each LLM can generate the correct answer and those where it cannot in a closed-book setting. Based on this division, we apply three rewriting methods and measure MRR and nDCG@10 scores as shown in Table~\ref{tab:closedbook}. Both the Mistral-7b and Phi-2 models demonstrate superior performance in the \textsc{Incorrect Answer} cases when using our rewriting method, compared to previous rewriting approaches. This demonstrates that our method achieves more effective information retrieval when the model needs to search unknown information, aligning with the original purpose of using RAG~\citep{10.5555/3495724.3496517} which is to use external knowledge when QA systems do not have knowledge to generate correct answer. In this experiment, we use the Contriever~\citep{izacard2022unsupervised} model for the dense retriever. 
\begin{table}[t]
\small

\renewcommand{\arraystretch}{1.4}
\resizebox{\linewidth}{!}{
\begin{tabular}{l|ccc|ccc}
\hline
{HotpotQA} & \multicolumn{3}{c}{{Correct Answer}}& \multicolumn{3}{c}{{Incorrect Answer}}  \\ \hline
% Mistral-7b& {Ours}& {Q2D}&{Q2C}& {Ours}& {Q2D}& {Q2C}    \\
% {nDCG@10} & {53.57}& {52.21} & {\textbf{54.95}} & {\textbf{49.61}}  & {39.72}  & {41.28}  \\
% {MRR}                                  & {\textbf{67.47}}          & {62.56} & {66.65} & {\textbf{58.69}}  & {49.68}  & {49.30}  \\ \hline
% Phi-2                                & {Ours}           & {Q2D}   & {Q2C}            & {Ours}            & {Q2D}    & {Q2C}    \\
% {nDCG@10}                              & {53.51} & {49.57} & {\textbf{55.36}} & {\textbf{50.32}}  & {37.60}  & {37.73}  \\
% {MRR}                                  & {\textbf{66.19}} & {59.90} & {63.96}          & {\textbf{57.04}}  & {47.27}  & {44.39}  \\ \hline
{Mistral-7b}& {Ours}& {Q2D}&{Q2C}& {Ours}& {Q2D}& {Q2C}    \\
{nDCG@10} & {77.80}& {76.00} & {\textbf{79.27}} & {\textbf{60.05}}  & {55.57}  & {58.77}  \\
{MRR}& {90.04}& {87.73} & {\textbf{90.90}} & {\textbf{78.76}}  &{71.71}  & {75.35}  \\ \hline
{Phi-2}& {Ours}& {Q2D}& {Q2C}& {Ours}& {Q2D}& {Q2C}    \\
{nDCG@10}& {74.72} & {71.03} & {\textbf{75.43}} & {\textbf{59.42}}  & {46.93}  & {57.64}  \\
{MRR}& {\textbf{87.95}} & {86.17} & {87.75}& {\textbf{79.16}}  & {64.63}  & {75.59}  \\ \hline
\end{tabular}
}
\caption{The impact of reliance on rewriting model internal knowledge.}
% \vspace{-4mm}
\label{tab:closedbook}
\end{table}

\paragraph{Evaluating on Open-Domain QA}
To evaluate the answer generation results based on retrieved passages, we conducted performance measurements using HotpotQA~\citep{yang2018hotpotqa} and NaturalQA~\citep{47761} datasets. We performed retrieval using rewritten queries, altering the number of passages retrieved with three different rewriting methods to measure accuracy as follows:
\begin{equation}
    q^{\prime} = \mathcal{M}_{\text{rewriting}} (\text{prompt} (q))
    \label{eq:rewrite_q}
\end{equation}
\begin{equation}
    p_1,\ldots, p_k = \mathcal{M}_{\text{Retriever}} (q^{\prime})
    \label{eq:retrieve}
\end{equation}
\begin{equation}
    \text{output}=\mathcal{M}_{\text{Generator}}(\text{prompt}(q, p_1, ..., p_k))
    \label{eq:generation}
\end{equation}
Both the generation and rewriting models utilized the Mistral-7b~\citep{jiang2023mistral} model, while the Contriever~\citep{izacard2022unsupervised} model served as the retrieval model. In Eq~\ref{eq:rewrite_q} and~\ref{eq:retrieve}, $q$ is the original query, and $q^{\prime}$ is the rewritten query, while $p_1, ..., p_k$ represents the retrieved passages, and the retriever searches the top k most relevant passages. We generate the final answer based on the retrieved passages and the query as shown in Eq~\ref{eq:generation}. For more details on our prompts refer to the Appendix~\ref{appen:prompt}. Figure~\ref{fig:qa} illustrates that our rewriting approach yields the best performance across both HotpotQA and NaturalQA. Additionally, we observed that the Q2D method's performance decreases when the number of passages ($k$) reaches seven. However, the \name{} method demonstrates superior accuracy compared to the Q2D and Q2C methods. Our method performs better in situations where the rewriting model generates incorrect answers, making it effective for open-domain QA. We conduct an analysis of this in Tables~\ref{tab:closedbook} and~\ref{tab:closedbook_ans_modi}.
% As mentioned in Tables~\ref{tab:closedbook} and~\ref{tab:closedbook_ans_modi}, our method performs effectively because it retrieves relevant passages better for information that is not known in the answers.

\begin{table}[t]
\small
\begin{tabular}{lcc}
\hline
\multicolumn{1}{c}{}& \multicolumn{2}{c}{HotpotQA}\\
\multicolumn{1}{c}{\multirow{-2}{*}{Mistral-7b}} & nDCG@10& MRR\\ \hline
New Query                 & \multicolumn{1}{l}{} & \multicolumn{1}{l}{} \\
\quad \name{}& \textbf{65.49}& \textbf{82.21}\\
\quad query2doc (Q2D)& 61.83& 76.62\\
\quad query2cot (Q2C)& 65.04& 80.11\\
Replace Answer to [MASK]&&\\
\quad \name{}& \textbf{64.09}& \textbf{81.45}\\
\quad query2doc (Q2D)& 61.08& 76.36\\
\quad query2cot (Q2C)& 63.92& 79.59\\
Delete New Queries with Answer&&\\ 
\quad \name{}& \textbf{58.29}& \textbf{77.46}\\
\quad query2doc (Q2D)& 53.39& 69.91\\
\quad query2cot (Q2C)& 56.54& 73.68\\
\hline
\end{tabular}
\caption{The results of an answer modification experiment.}
\vspace{-4mm}
\label{tab:closedbook_ans_modi}
\end{table}
\paragraph{Experiment on the Impact of Answer Presence}
We examine the changes in retrieval performance based on the presence or absence of answers. In Table~\ref{tab:closedbook_ans_modi}, rewriting the HotpotQA dev dataset~\citep{yang2018hotpotqa}, comprising 5,447 entries using Mistral-7b, results in new queries that include 1,746, 1,623, and 1,752 answers for \name{}, Q2D, and Q2C, respectively. To evaluate the impact of our model's reliance on internal knowledge for generating answers, we conducted three experiments. First, in the \textit{New Query} setting, we directly utilize the new queries generated by the LLM. Second, in the \textit{Replace Answer to [MASK]} setting, if the new query contains an answer, we mask the answer portion. Finally, in the \textit{Delete New Queries with Answer} experiment, we exclude any new queries from the evaluation if at least one of the three rewriting methods generated a new query containing an answer. 
% as in Figure~\ref{fig:retr} dash line. 
The evaluation dataset reduces to 3,212 in \textit{Delete New Queries with Answer} experiment. Our approach achieves the best performance across all three experiments. We observe that replacing answers with \textit{[MASK]} results in a decrease in MRR and nDCG@10 performance across all rewriting methods. Notably, the difference in nDCG@10 scores between our method and Q2C was 0.45 in the first experiment and 1.75 in the third experiment. This score gap suggests that our approach is relatively less affected by the presence or absence of answers in the queries compared to existing methods.
 
% These new queries can serve as the final answer without the need for a retrieval process. To adhere to the original purpose of retriever as discussed in Section~\ref{lab:intro}. we conduct two experiments to assess the performance of each rewriting method. In the first experiment, we mask the answer segments; in the second, we exclude any data where the answer appears even once in the sections \name{}, Q2D, or Q2C. The evaluation dataset for the second experiment reduces to 3,212 in \textsc{Delete New Queries with Answer}. Our approach consistently outperforms existing methods across all experimental settings. Notably, within the \textsc{Delete New Queries with Answer}, the substantial performance improvement over existing methods indicates that our approach reduces dependency on the presence of answers. Additionally, the Contriever model is employed in this experiment.
\begin{table}[t]
\centering

\small
\renewcommand{\arraystretch}{1.2}

\resizebox{\linewidth}{!}{
\begin{tabular}{lccc}

\hline
\multicolumn{1}{c}{\multirow{2}{*}{BEIR}} & \multicolumn{3}{c}{FActScore}  \\
\multicolumn{1}{c}{}                              & Crafting the Path           & Q2D   & Q2C   \\ \hline 
Mistral-7b                                        & \textbf{0.718} & 0.506 & 0.711 \\
Phi-2                                             & \textbf{0.765} & 0.460 & 0.675 \\ \hline
\end{tabular}
}
\caption{Average FActScore~\citep{min2023factscore} on each method.}
\vspace{-4mm}
\label{tab:factscore}

\end{table}

\paragraph{Evaluating the Factuality of Queries}
To determine the impact of factual errors occurring during query rewriting on retrieval performance, we use FActScore to measure the accuracy of Mistral-7b and Phi-2 across three rewriting methods. Unlike \citet{min2023factscore}, which uses atomic facts, we simply divide the content by sentence. For each separated sentence, we use a gold label passage as evidence to output as \texttt{True} or \texttt{False}. If the factuality evaluation results for three sentences of a rewritten query are \texttt{True}, \texttt{True}, \texttt{False}, we assign a score of 2/3. We calculate the average for the measured queries. In Table~\ref{tab:factscore}, \name{} exhibits the highest FActScore (the smallest factual error), demonstrating the best performance with both the Mistral-7b and Phi-2 models. We utilize GPT-3.5 Turbo~\citep{OpenAI_ChatGPT, openai2023gpt4} as the factuality evaluation model.

\begin{table*}[t!]
\label{tab:case}
\footnotesize  
\centering
\resizebox{\textwidth}{!}{
\renewcommand{\arraystretch}{0.8}
\begin{tabular}{cccc}
\toprule
\multicolumn{1}{p{.12\textwidth}}{\textbf{Case}} &
\multicolumn{1}{p{.12\textwidth}}{\textbf{Question}} & \multicolumn{1}{p{.4\textwidth}}{\textbf{Crafting the Path}} & \multicolumn{1}{p{.4\textwidth}}{\textbf{query2cot (Q2C)}} \\

\midrule
\multicolumn{1}{p{.12\textwidth}}{\textbf{Case 1:} \newline Effectiveness of generating None. }&
\multicolumn{1}{p{.12\textwidth}}{\textbf{Question:} \newline What is an rs in money? }&
\multicolumn{1}{p{.4\textwidth}}{\textbf{New Query:} The abbreviation "RS" does not have a standard meaning in the context of money or finance.To understand the meaning of "RS" in a given financial context, \textbf{additional context or clarification from the source is required}. \textbf{None.} \newline - \textbf{(MRR: 33.3)}} &
\multicolumn{1}{p{.4\textwidth}}{\textbf{New Query:} In finance, an \textcolor{red}{RS (Relative Strength) ratio is a technical indicator that compares the stock's performance} against a benchmark index or another stock. It measures the strength of a security's price action in relation to the market, helping investors identify potential outperformance or under performance. - \textbf{(MRR: 12.5)}} \\ \midrule \midrule

\multicolumn{1}{p{.12\textwidth}}{\textbf{Case 2:} \newline Errors in a previous step lead to issues in subsequent generation.}&
\multicolumn{1}{p{.12\textwidth}}{\textbf{Question:} \newline What is simplicheck? }&
\multicolumn{1}{p{.4\textwidth}}{\textbf{New Query:} Simplicheck is a term or concept that may represent a specific tool, process, or concept, \textbf{but without further context, its meaning remains unclear}. To understand what Simplicheck is, more information or context is required. \textbf{None.} \newline - \textbf{(MRR: 16.7)}} &
\multicolumn{1}{p{.4\textwidth}}{\textbf{New Query:} \textcolor{blue}{Simplicheck is a software tool} designed for automated vulnerability scanning and penetration testing. It helps identify security weaknesses in web applications, networks, and systems, enabling organizations to prioritize and remediate risks effectively. - \textbf{(MRR: 100)}} \\ \midrule \midrule

\multicolumn{1}{p{.12\textwidth}}{\textbf{Case 3:} \newline factual errors occur. }&
\multicolumn{1}{p{.12\textwidth}}{\textbf{Question:} \newline Tricare service number.}&
\multicolumn{1}{p{.4\textwidth}}{\textbf{New Query:} TRICARE is a health program for eligible active duty and retired military personnel, their families, and survivors. To access TRICARE services, members need to provide their unique identifier, such as a service number. he TRICARE service number is a \textcolor{red}{12-digit} identifier, typically \textcolor{red}{formatted as NNN-NNN-NNNN-NN-NN-NN}. - \textbf{(MRR: 11.11)}} &
\multicolumn{1}{p{.4\textwidth}}{\textbf{New Query:} TRICARE is a health program for active-duty military, retirees, and their families. To obtain specific information, contact TRICARE Customer Service at \textcolor{blue}{1-877-874-2273} (option 1) for enrollment, eligibility, and benefits inquiries. This number ensures access to accurate and timely information. - \textbf{(MRR: 20)}} \\

\bottomrule

\end{tabular}}
\caption{Case study of reformulated queries generated using Mistral-7b with Crafting the Path and Q2C. MRR represents the retrieval performance when using the new query. We represent the factual error in red and the accurate information in blue.}
% \vspace{-4mm}
\end{table*}

% \multicolumn{1}{p{.12\textwidth}}{\textbf{Case3:} factual errors occur. \newline \textbf{Question:} Tricare service number}&
% \multicolumn{1}{p{.4\textwidth}}{\textbf{Query Type: } A (Non Retrieval)\newline \textbf{Reasoning: }César Gaytan was born in the continent of \textcolor{blue}{North America}. The Italian navigator who explored the eastern coast of North America for the English was \textcolor{blue}{Giovanni Caboto}, also known as John Cabot. So the answer is: \textcolor{red}{Giovanni Caboto/John Cabot}.} &
% \multicolumn{1}{p{.4\textwidth}}{\textbf{Query Type: } C (Multi-step Approach)\newline \textbf{Reasoning: }Césarr Gaytan was born in \textcolor{blue}{Guadalajara, Jalisco, Mexico}. The Italian navigator who explored the eastern coast of the continent for the English is \textcolor{blue}{John Cabot}. John Cabot's son is \textcolor{blue}{Sebastian Cabot}. So the answer is: \textcolor{blue}{Sebastian Cabot.}} \\

\begin{figure}[ht]
\centerline{\includegraphics[scale=0.35]{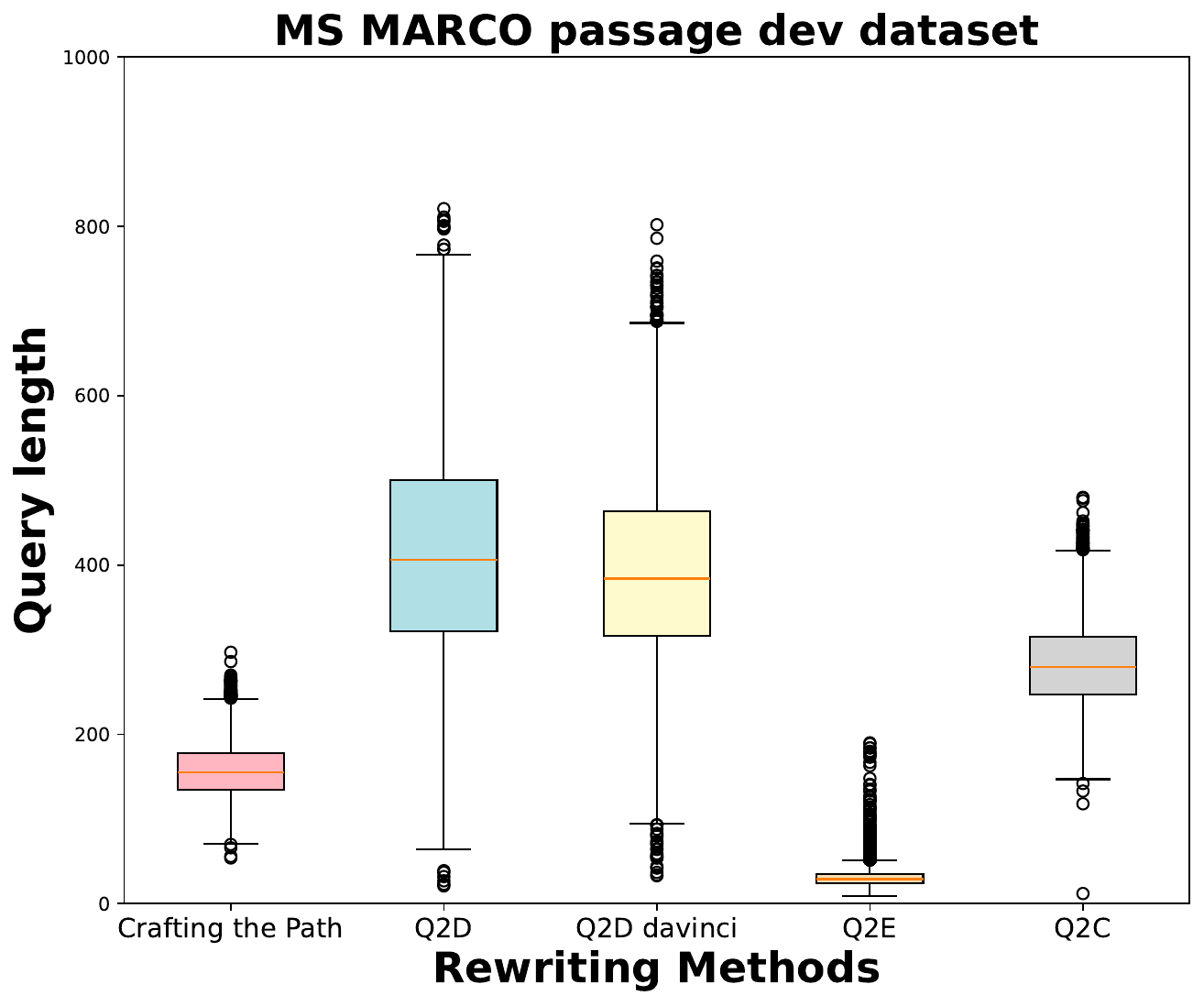}}
\caption{The length of the new query for each rewriting method in the MS-MARCO passage dev dataset.}
\vspace{-6mm}
\label{fig:ds_len}
\end{figure}
\begin{table}[ht]
\centering
\small
\renewcommand{\arraystretch}{1.5}
\resizebox{\linewidth}{!}{
\begin{tabular}{lcc}

\hline
                  & LLM call & Index search \\ \hline
\name{} & 5648.9ms &  261.44ms  \\
query2doc (Q2D)         & 8167.5ms   &     335.65ms    \\
query2cot (Q2C)         & 6094.2ms   &     281.98ms    \\
 \hline
\end{tabular}
}
\caption{Latency analysis of each method in BM25 search and LLM call from the MS-MARCO passage dev dataset.}
\vspace{-4mm}
\label{tab:latency}
\end{table} 

\paragraph{Query Length and Latency}
\label{sec:latency}
In Figure~\ref{fig:ds_len}, we compare the query length for each rewriting method. Additionally, we include the prompts utilized for each rewriting method in Appendix~\ref{appen:prompt}. 
% In Figure~\ref{fig:ds_len}, we measure the string length of queries to observe the impact of rewritten query length on the model's generation latency. Q2D davinci\footnote{https://huggingface.co/datasets/intfloat/query2doc\_msmarco} which contains GPT-3.5 (text-davinci-003) generations measures the query length based on the queries generated by~\citet{wang2023query2doc}. Additionally, we generate results for other methodologies based on the prompts provided in the Appendix~\ref{appen:prompt}. 
We observe that the \name{} approach has a shorter context length than all other methods except for Q2E. Our approach maintains a structured format, which allows us to minimize unnecessary word generation while achieving superior performance. In Table~\ref{tab:latency}, since all three methods use the same dense retriever architecture, the search time is equal. Therefore, we measure the search time using the sparse retriever (BM25) instead. Hence, we measure the time of LLM call latency to compare the speed of each method. We retrieve results from the MS-MARCO passage dev dataset~\citep{Campos2016MSMA} for 1,000 entries and average the outcomes over 100 repetitions. We measure the latency incurred when the model generates a new query and BM25 searches for relevant passages. Our method generates less redundant information, resulting in lower latency compared to Q2D. Also, our method shows comparable latency to Q2C because both methods have a low context length.

\paragraph{Case Study}
In Table 8, we conduct a representative case analysis of both the strengths and limitations of our proposed method. Case 1 shows an improvement in performance by minimizing factual errors by generating "None". In contrast, we observe factual errors occurring in Q2C during the reasoning process for answer generation. The retrieval performance of Q2C decreases compared to \name{} due to these factual errors. Conversely, our method generates "None" for unknown contexts, minimizing factual errors and improving retrieval performance. In Case 1, RS is the currency of India. Q2C generates an explanation of the RSI performance indicator. In Case 2, We demonstrate that errors in a previous step propagate in subsequent generations. The \name{} method fails to provide a correct explanation of Simplicheck in step 1, the Query Concept Comprehension process. In subsequent steps, it indicates that more information is needed. Also, as in Case 3, our method does not completely eliminate factual errors. In the case of Q2C, it generates the correct number, resulting in better retrieval performance. On the other hand, our method generates the wrong number format. Additionally, the proportion of queries that generated "None" among all rewritten queries is 3.3\%.

% Our method can affect steps 2 and 3 when incorrect information generation in the QCC stage (step1) affects subsequent steps.
\section{Related Work}
\paragraph{Information Retrieval}
Information retrieval is the process of obtaining relevant information from the database based on given queries. The main two methods for information retrieval are sparse retrieval and dense retrieval. A prominent example of the sparse retrieval method is BM25~\citep{robertson1976relevance, robertson1995okapi}, which serves as a ranking function to evaluate the relevance between a given query and documents. In contrast, dense retrieval method~\citep{xiong2020approximate,Qu2021RocketQAAO} involves fetching passages that exhibit high similarity to the query using the document embeddings. This approach typically utilizes pre-trained language models such as BERT~\citep{devlin2018bert} for the encoder, and some methods fine-tune these encoders.~\citep{Karpukhin2020DensePR, Wang2022SimLMPW}. 
%In sparse retriever, five original queries and the rewritten query are concatenated and used in the retrieval system. For dense retriever, one original query and the rewritten query are interleaved with a [SEP] token for separating~\citep{wang2023query2doc}. 
%Recent research has also explored generative retrieval methods~\citep{tay2022transformer, sun2024learning}, employing encoder-decoder models to encode document within model parameters. 
In this work, we improve the performance of both sparse and dense retrieval methods by focusing on query rewriting method.
%In this approach, the model receives a query as input and outputs the corresponding document.

%\paragraph{Pseudo-Relevance Feedback (PRF)}
%In classic methods, PRF~\citep{Lavrenko2001RelevanceBasedLM} exists. PRF treats documents retrieved by the original query as "pseudo-relevant" and uses them to extract new query terms. However, PRF based approaches assume that the top-k retrieved documents are relevant to the query. This assumption worsens the problem when the query is short or ambiguous, leading to retrieved documents not perfectly aligning with the original query. This misalignment can result in poor outcomes in the final results.
% \begin{figure*}[!ht]
% \centerline{\includegraphics[scale=0.50]{./images/figure2.pdf}}
% % \vspace{-4mm}
% \caption{Overview of our proposed query rewriting method \name{}, along with the rewritten query examples of \textit{query2doc} (Q2D) and \textit{query2cot} (Q2C) methodologies. We represent the factual error in red and the accurate information in blue.}
% \vspace{-5mm}
% \label{fig:crafting}
% \end{figure*}
\paragraph{LLM Based Query Rewriting}
Query rewriting refers to the task that modifies the original query to improve the search results for the information retrieval systems. Some studies on query rewriting have employed neural networks to produce or select expansion terms~\citep{zheng2021contextualized, roy2016using, imani2019deep}, typically through training or fine-tuning a model. In contrast, our approach leverages the inherent capabilities of Large Language Models (LLMs) without requiring training or fine-tuning. Recent studies on query rewriting mainly use the large language models to create relevant information for the given query. \textit{query2doc} (Q2D)~\citep{wang2023query2doc} operates by generating a pseudo-document based on the original query, which is then used as input for retriever. Similarly, \textit{query2cot} (Q2C) employs a Chain of Thought (CoT)~\citep{wei2022chain} approach, and \textit{query2expand} (Q2E)~\citep{jagerman2023query} generates semantically equivalent query. These approaches leverage the rewriting of the original query into a form similar to passages in the corpus. This technique leads to significant performance improvement compared to using the base query alone.  
Furthermore, \textit{Rewrite-Retrieve-Read}\citep{ma-etal-2023-query} introduces a methodology that enhances rewriting performance by incorporating reinforcement learning. Another study, \textit{ITER-REGEN}\citep{shao-etal-2023-enhancing}, improves query quality by feeding the query and retrieved documents into a language model for rewriting, followed by a repeated retrieval process. \textit{Rephrase and Respond}~\citep{deng2023rephrase} argues that for effective rewriting, queries should be rephrased in a manner that is easier for LLMs to understand. For the conversational serach, \citep{yoon2024ask} proposes a method that generates a variety of queries and uses the rank of retrieved passages to train the LLMs on only the optimal queries. This process is further refined using a DPO~\citep{rafailov2023direct} approach to create optimal queries.
%Our work aligns similarities to recent efforts such as Q2D and Q2E in terms of leveraging linguistic techniques to expand queries. However, unlike previous approaches, we focus on generating only the necessary information to retrieve for the query and aim to enhance performance by mitigating hallucination in the reformulated query.

Our work has similarities with recent efforts like Q2D~\citep{wang2023query2doc} and Q2E~\citep{jagerman2023query}, particularly in using linguistic techniques to expand queries. Q2D and Q2C perform well on queries where the rewriting model can produce accurate answers, but their performance significantly deteriorates when inaccurate answers occur. However, unlike previous approaches, we focus on rewriting queries to improve search performance when the model generates incorrect answers. We demonstrate significant improvements in QA tasks through this problem mitigation. Additionally, we aim to minimize the generation of inaccurate information and concentrate on producing the necessary data for information retrieval.

\section{Conclusion}
We present a Crafting the Path, an approach that involves a structured three-step process, focusing not merely on generating additional information for the query but primarily on generating what information to find from the query.
Our approach shows superior retrieval performance compared to the existing rewriting method, achieves improved performance in open domain QA by being less reliant on internal model knowledge and demonstrates robust performance across various models. Additionally, the method generates fewer factual errors and delivers improved out-of-domain performance with lower latency than previous methods. 

Tasks such as query rewriting in information retrieval can benefit from the advancements in LLMs. Additionally, as LLMs become more universally accessible, they can become a central part of information retrieval systems. This offers the advantage of providing users with more accurate information.
%To address the challenges associated with the dependence on internal parameter knowledge characteristic of previous rewriting models, we propose the Crafting the Path method. 
% Our method less relies on internal model parameters and demonstrates robust performance across various models. Additionally, the method generates fewer factual errors and delivers improved out-of-domain performance with lower latency than previous methods.
\section*{Limitations}
Our method outperforms existing ones, but using an LLM for query rewriting inherently introduces latency. However, we propose a rewriting method that, compared to existing methods, results in relatively lower latency while offering better retrieval performance. Additionally, we experimentally demonstrate scenarios where rewriting can and cannot generate answers. However, we do not present an automatic method to distinguish between these scenarios in actual applications. We leave the study of such methods for future research.
\section*{Ethics Statement}
This study conducts query rewriting and QA tasks using an LLM, and also searches for relevant documents. Since we carry out generation tasks based on the LLM, it is important to be aware that the LLM may produce inappropriate responses. Additionally, as it may retrieve inappropriate content from the searched documents, developing management methods for this is essential. We believe this is a crucial area for future work.

\section*{Acknowledgement}
This research was supported by Institute for Information \& Communications Technology Planning \& Evaluation (IITP) through the Korea government (MSIT) under Grant No. 2021-0-01341 (Artificial Intelligence Graduate School Program (Chung-Ang University)).

\bibliography{anthology,custom}

\appendix

\appendix
\newpage
\clearpage
\section{Appendix}
\subsection{Hyperparameters}
We present the detailed hyperparameters in Table~\ref{tb:biencoder-hyper-params}.

\label{appen:dense_hyper}
\begin{table}[hbt]
  \centering
  \small{
    \begin{tabular}{l|c}
      \toprule
      \textbf{Name}                    & \textbf{Value}  \\
      \midrule
      Learning rate & 2e-5    \\
      PLM  & all-mpnet-base-v2    \\
      Batch & 128 \\
      Epoch      & 3   \\
      Learning rate decay     & linear \\
      Warmup steps            & 1000   \\
      Binary loss margin                 & 2.0    \\
      Similarity function            & dot score    \\
      Query length         & 128    \\
      Passage length          & 128  \\
      \bottomrule
    \end{tabular}
  }
  \caption{Hyperparameters used to train dense retrieval model.}
\label{tb:biencoder-hyper-params}
\end{table}
\subsection{Prompts}
\label{appen:prompt}

\begin{table*}[ht]
\centering
\scalebox{0.85}{\begin{tabular}{ll}
\hline
prompts        & \begin{tabular}[c]{@{}p{1.0\linewidth}@{}}\textbf{Instruction:}
Based on the example below, write 3 steps related to the Query and answer in the same format as the example.\\
\\
\textbf{Requirements:}\\
1. In step1, sub-information from the existing query is extracted.\\
2. In step2, please generate what information is needed to solve the question.\\
3. In step3, an answer is generated based on Query, step1, and step2.\\
4. If you don't have certain information, generate 'None'.\\
5. Please prioritize your most confident predictions.\\
\\
\textbf{Example:}\\
\textbf{Query:} where is the Danube?\\
\textbf{step1:} The Danube is Europe's second-longest river, flowing through Central and Eastern Europe, from Germany to the Black Sea.\\
\textbf{step2:} To locate the Danube precisely, geographical knowledge or a map of Europe highlighting rivers is necessary.\\
\textbf{step3:} The Danube flows through 10 countries.\\
\\
\textbf{Query:} what is the number one formula one car?\\
\textbf{step1:} Formula One (F1) is the highest class of international automobile racing competition held by the FIA.\\
\textbf{step2:} To know the best car, you have to look at the race records.\\
\textbf{step3:} Red Bull Racing's RB20 is the best car. \\
\\
\textbf{Query:} which movie did Michael Winder write?\\
\textbf{step1:} Michael Winder is a screenwriter involved in the film industry, potentially credited with writing one or more movies.\\
\textbf{step2:} To identify the movie(s) Michael Winder wrote, access to a film database or filmography reference is needed.\\
\textbf{step3:} Michael Winder wrote the movie "In Time" (2011).\\
\\
\textbf{Query:} who's the director of Predators?\\
\textbf{step1:} "Predators" is a film, and like all films, it has a director responsible for overseeing the creative aspects of the production.\\
\textbf{step2:} To identify the director of "Predators," one needs access to movie databases, film credits, or industry knowledge about this specific film.\\
\textbf{step3:} Nimród Antal is the director of "Predators" (2010).\\
\\
\textbf{Query:}\\
\end{tabular}  \\ \hline
\end{tabular}}
\caption{The full prompt used for \name{} method.}
\label{tab:prompt crafting}
\end{table*}

%Figure ~\ref{fig:few_shot}

\begin{table*}[ht]
\centering
\scalebox{0.85}{\begin{tabular}{ll}
\hline
prompts        & \begin{tabular}[c]{@{}p{1.0\linewidth}@{}}\textbf{Instruction:}\\
You are good at writing Passage. You are asked to write a passage that answers the given query. Do not ask the user for further clarification.\\
\\
\textbf{Requirements:}\\
1. Please write it in a similar format to the example\\
2. Please prioritize your most confident predictions.\\
\\
\textbf{Example:}\\
\textbf{Query:} what state is this zip code 85282\\
\textbf{Passage:} Welcome to TEMPE, AZ 85282. 85282 is a rural zip code in Tempe, Arizona. The population\\
is primarily white, and mostly single. At \$200,200 the average home value here is a bit higher than\\
average for the Phoenix-Mesa-Scottsdale metro area, so this probably isn’t the place to look for housing\\
bargains.5282 Zip code is located in the Mountain time zone at 33 degrees latitude (Fun Fact: this is the\\
same latitude as Damascus, Syria!) and -112 degrees longitude.\\
\\
\textbf{Query:} why is gibbs model of reflection good\\
\textbf{Passage:} In this reflection, I am going to use Gibbs (1988) Reflective Cycle. This model is a recognised\\
framework for my reflection. Gibbs (1988) consists of six stages to complete one cycle which is able\\
to improve my nursing practice continuously and learning from the experience for better practice in the\\
future.n conclusion of my reflective assignment, I mention the model that I chose, Gibbs (1988) Reflective\\
Cycle as my framework of my reflective. I state the reasons why I am choosing the model as well as some\\
discussion on the important of doing reflection in nursing practice.\\
\\
\textbf{Query:} what does a thousand pardons means\\
\textbf{Passage:} Oh, that’s all right, that’s all right, give us a rest; never mind about the direction, hang the\\
direction - I beg pardon, I beg a thousand pardons, I am not well to-day; pay no attention when I soliloquize,\\
it is an old habit, an old, bad habit, and hard to get rid of when one’s digestion is all disordered with eating\\
food that was raised forever and ever before he was born; good land! a man can’t keep his functions\\
regular on spring chickens thirteen hundred years old.\\
\\
\textbf{Query:} what is a macro warning\\
\textbf{Passage:} Macro virus warning appears when no macros exist in the file in Word. When you open\\
a Microsoft Word 2002 document or template, you may receive the following macro virus warning,\\
even though the document or template does not contain macros: C:\textbackslash<path>\textbackslash<file name>contains macros.\\
Macros may contain viruses.\\
\\
\textbf{Query:}
\end{tabular}  \\ \hline
\end{tabular}}
\caption{The full prompt used for \textit{query2doc} (Q2D) method.}
\label{tab:prompt q2d}
\end{table*}

\begin{table*}[ht]
\centering
\scalebox{0.85}{\begin{tabular}{ll}
\hline
prompts        & \begin{tabular}[c]{@{}p{1.0\linewidth}@{}}\textbf{Instruction:}\\
Based on the example below, write keywords. Do not ask the user for further clarification\\
\\
\textbf{Requirements:}\\
1. Please write it in a similar format to the example\\
2. Please prioritize your most confident predictions.\\
\\
\textbf{Example:}\\
\textbf{Query:} how to include bullets in excel\\
\textbf{Keywords:} insert bullet points in excel\\
\\
\textbf{Query:} positive predictive value formula\\
\textbf{Keywords:} calculating positive predictive value\\
\\
\textbf{Query:} house for sale bridgewater ma\\
\textbf{Keywords:} homes for sale in bridgewater\\
\\
\textbf{Query:} r text command\\
\textbf{Keywords:} text processing in r\\
\\
\textbf{Query:}\\
\end{tabular}  \\ \hline
\end{tabular}}
\caption{The full prompt used for \textit{query2expand} (Q2E) method.}
\label{tab:prompt q2e}
\end{table*}

%Figure ~\ref{fig:few_shot}

\begin{table*}[ht]
\centering
\scalebox{0.85}{\begin{tabular}{ll}
\hline
prompts        & \begin{tabular}[c]{@{}p{1.0\linewidth}@{}}\textbf{Instruction:}\\
Answer the following query. Give the rationale before answering:\\
\\
\textbf{Requirements:}\\
1. Please write it in a similar format to the example\\
2. Please prioritize your most confident predictions.\\
3. Let's think step by step. \\
\\
\textbf{Query:} what does folic acid do\\
\textbf{Answer:} Folic acid aids in DNA synthesis, cell division, and red blood cell formation. It's vital for fetal development during pregnancy, preventing neural tube defects, and supporting general health.\\
\\
\textbf{Query:} what is calomel powder used for?\\
\textbf{Answer:} Calomel powder, historically used in medicine, served as a purgative, diuretic, and syphilis treatment. Its usage declined due to the toxic effects of mercury, leading to safer alternatives. Today, it's largely obsolete in medical practice.\\
\\
\textbf{Query:} what county is dewitt michigan in?\\
\textbf{Answer:} DeWitt, Michigan, is located in Clinton County. This geographic classification helps in understanding local governance, services, and regional affiliations, essential for residents and researchers.\\
\\
\textbf{Query:} the importance of minerals in diet\\
\textbf{Answer:} Minerals are crucial for bodily functions, including bone health, fluid balance, and muscle function. They support metabolic processes and the nervous system, highlighting their essential role in maintaining overall health and preventing deficiencies.\\
\\
\textbf{Query:}
\end{tabular}  \\ \hline
\end{tabular}}
\caption{The full prompt used for \textit{query2cot} (Q2C) method.}
\label{tab:prompt cot}
\end{table*}

\newpage
\clearpage

\end{document}